%% file: main.tex
\pdfoutput=1
\documentclass[conference]{IEEEtran}
\include{preamble}

\begin{document}

\title{
 A Minmax Utilization Algorithm for Network Traffic Scheduling of Industrial Robots
    \thanks{This research is partially funded by the Innovate UK Automation of Network edge Infrastructure \& Applications with aRtificiAl intelligence ukANIARA project, which is part of the EU Celtic Next project ANIARA (\url{www.celticnext.eu/project-ai-net-aniara}).}
}

\author{
    \IEEEauthorblockN{Yantong Wang\IEEEauthorrefmark{1}\IEEEauthorrefmark{2},  Vasilis Friderikos\IEEEauthorrefmark{2}, Sebastian Andraos\IEEEauthorrefmark{3}}
    \IEEEauthorblockA{\IEEEauthorrefmark{1} School of Information Science and Engineering, Shandong Normal University, Ji'nan, 250358, China}
    \IEEEauthorblockA{\IEEEauthorrefmark{2} Department of Engineering, King's College London,
    London, WC2R 2LS, U.K.}
    \IEEEauthorblockA{\IEEEauthorrefmark{3} HAL Robotics Ltd, 115 Coventry Road, London, E2 6GG, U.K.}
    \IEEEauthorblockA{\{yantong.wang, vasilis.friderikos\}@kcl.ac.uk, s.andraos@hal-robotics.com}
}

\maketitle

\begin{abstract}
    Emerging 5G and beyond wireless industrial virtualized networks are expected to support a significant number of robotic manipulators. Depending on the processes involved, these industrial robots might result in significant volume of multi-modal traffic that will need to traverse the network all the way to the (public/private) edge cloud, where advanced processing, control and service orchestration will be taking place. In this paper, we perform the traffic engineering by capitalizing on the underlying pseudo-deterministic nature of the repetitive processes of robotic manipulators in an industrial environment and propose an integer linear programming (ILP) model to minimize the maximum aggregate traffic in the network. The task sequence and time gap requirements are also considered in the proposed model. To tackle the curse of dimensionality in ILP, we provide a random search algorithm with quadratic time complexity. Numerical investigations reveal that the proposed scheme can reduce the peak data rate up to $53.4\%$ compared with the nominal case where robotic manipulators operate in an uncoordinated fashion, resulting in significant improvement in the utilization of the underlying network resources.
\end{abstract}
\begin{IEEEkeywords}
Integer Linear Programming, Industrial Robots, Network Traffic Scheduling, Network Optimization, Industry 4.0
\end{IEEEkeywords}

\input{section/introduction}
\input{section/motivation}
\input{section/model}
\input{section/heuristic}

\input{section/numerical}

\input{section/conclusions}

\bibliographystyle{ieeetr}
\bibliography{reference}

\end{document}

%% file: preamble.tex
\IEEEoverridecommandlockouts
\usepackage[english]{babel}
\usepackage[utf8x]{inputenc}
\usepackage{booktabs}
\usepackage{cite}
\usepackage{graphicx}
\usepackage{amsmath,amssymb,amsfonts}
\usepackage[ruled,linesnumbered]{algorithm2e}
\usepackage{xcolor}
\usepackage{textcomp}
\usepackage{url}
\usepackage{hyperref}
\usepackage{caption}
\usepackage{setspace}
\usepackage{subcaption}
\usepackage{multirow}
\usepackage{multicol}
\usepackage{comment}
\usepackage{float}

\DeclareMathAlphabet\mathcal{OMS}{cmsy}{m}{n}
\SetMathAlphabet\mathcal{bold}{OMS}{cmsy}{b}{n}

\def\BibTeX{{\rm B\kern-.05em{\sc i\kern-.025em b}\kern-.08em
    T\kern-.1667em\lower.7ex\hbox{E}\kern-.125emX}}
    
\allowdisplaybreaks

\newcommand\figSize{0.48}
\newcommand\figScale{0.17}

%% file: section/introduction.tex
\section{Introduction}
\label{sec:intro}

\IEEEPARstart{A}{s} global economic recovery continues amid a resurging Covid-19 pandemic, manufacturing is accelerating Industry 4.0 and Smart Factory deployment as a way to reduce human-to-human interactions. To this end, automated industrial robots are widely emerging on assembly lines to, inter alia, reduce infection risk for human workers and ensure continuous production. Therefore, it is not surprising that very labour-intensive tasks are now shifting towards being automated. The advent of edge cloud supported 5th Generation (5G) technology that enables untethered plug-and-play operation propels a high degree of efficient and flexible customization for robotic manipulators in production lines~\cite{rao2018impact} . As part of the 5G standardization the so-called Ultra Reliable Low Latency Communication (URLLC) use cases are specifically designed for enabling Machine-type Communication (MTC) systems including factory automation and robotic manipulators \cite{3GPP_1}.
The role of edge cloud based 5G networks in industrial settings together with architectural perspectives, use case scenarios and the issue of network slicing in such setting are detailed in \cite{aleksy2019utilizing}, \cite{Netslice}. In this emerging ecosystem the central aim of this work, as eluded in a more detailed manner hereafter, is to provide advanced traffic engineering by capitalizing on the pseudo-deterministic nature of tasks carried out by industrial robot manipulators.

Optimization for industrial robots has attracted significant attention in prior works~\cite{touzani2021multi,mazdin2021distributed,subha2021distributed,singh2021adaptive,li2015subtask,chen2018qos}. In order to solve the multi-task sequencing problem in an automotive assembly line, the authors in~\cite{touzani2021multi} develop a genetic algorithm-based approach that minimizes the production time and the duration of robots' motion. The study in~\cite{mazdin2021distributed} considers multi-robot scenarios with possible communication failure and  proposes distributed coalition formation and task assignment methods. In~\cite{subha2021distributed}, the authors regard the multi-robot system as a non-linear system and propose a distributed leader-follower adaptive consensus method to deal with time-varying faults. Additionally, the work in~\cite{singh2021adaptive} designs an energy-efficient communication strategy in multi-robot systems by employing the flower pollination algorithm. 
In~\cite{li2015subtask}, the authors focus on the cloud manufacturing platform scheduling and allocate distributed robots to process a batch of tasks simultaneously. The study in~\cite{chen2018qos} designs a computation offloading method with the consideration of quality-of-service for cloud robotics. However, the majority of the above works do not consider the network resource, especially the bandwidth, when designing the allocation strategies~\cite{touzani2021multi,mazdin2021distributed,subha2021distributed,singh2021adaptive,li2015subtask}. Moreover, the research in~\cite{chen2018qos} employs the bandwidth utilization as an energy indicator but ignores the network resource limitations.

There is also research considering the bandwidth constraint in the traffic control~\cite{kim2018xmas,dai2019statistical,xiao2019sensor,cui2020tclivi,vlk2020enhancing,zhao2018improving,specht2017synthesis}. The authors in~\cite{kim2018xmas} propose a scheme called XMAS to improve the accuracy of bandwidth estimation and determine the transmission bitrate. In~\cite{dai2019statistical}, a linear-regression learning model is used for sending data rate adjusting. Moreover, two adaptive bitrate algorithms, which are based on deep reinforcement learning model, are provided in~\cite{xiao2019sensor} and~\cite{cui2020tclivi} for unmanned aerial vehicle transmission and multimedia broadcast scenarios respectively. In time-sensitive networking, various traffic shapers are introduced for data flow control, such as time-aware shaper~\cite{vlk2020enhancing}, credit-based shaper~\cite{zhao2018improving}, and asynchronous traffic shaper~\cite{specht2017synthesis}. Though the aforementioned algorithms can improve the network performance under various scenarios, they can not be mitigated to the traffic control of industrial robot systems directly  since these works do not take the order of the task sequence into consideration. 


In this paper, the main contribution is to propose an integer linear programming (ILP) model for industrial robots to minimize peak data rate and  thereby reduce the bandwidth requirement. The task sequence order and time gap requirements are also considered. To accelerate the solution, we also provide a randomized search algorithm with quadratic time complexity. The numerical investigations reveal that the proposed ILP and the heuristic algorithm can save around $40\%$ and $30\%$ peak data rate compared with the nominal case of non coordinated traffic scheduling. 
The existing research in~\cite{el2016environment} is closest to the problem we tackle here. Though~\cite{el2016environment} also considers the reduction of bandwidth usage and the collaborations among robots, there are some main differences between our paper and~\cite{el2016environment}. On the one hand, the modelling granularity are different. In this paper, the communication process is divided into several sub-task sequences and we optimize the bandwidth usage by controlling the behavior of each sub-task; while in~\cite{el2016environment}, the communication task is treated entirely. On the other hand, the work in~\cite{el2016environment} optimize the traffic data rate by allocating each channel bandwidth, which is inappropriate for some jitter-sensitive applications such as the example in Section~\ref{sec:motiv}. In this paper, we orchestrate the sub-tasks' processing time slot to provide a smooth communication. 

%% file: section/motivation.tex
\section{Motivation and Industrial Example}
\label{sec:motiv}
To motivate the research, we use the following instance to illustrate the traffic scheduling issue in industrial productions.

\begin{figure}[htbp]
    \centering
    \includegraphics[width=.48\textwidth]{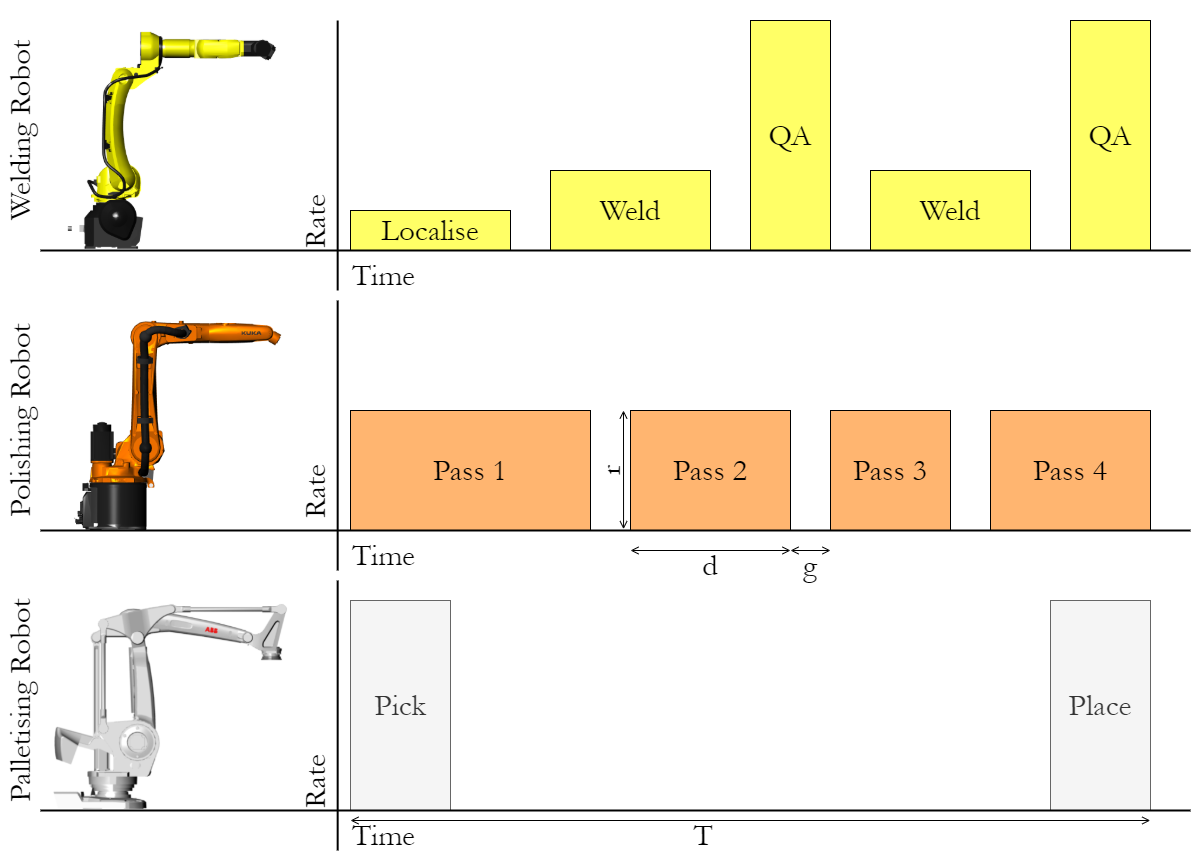}
    \caption{Example tasks with relative durations and data rates.}
    \label{fig:tasks}
\end{figure}

The tasks undertaken by robots can be highly varied, in both process and transport requirement, within the same facility. 
As shown in Figure~\ref{fig:tasks}, one cell\footnote{A cell is the combination of robots, tooling and sensors which work together to undertake a given task.} could be dedicated to welding parts, another to polishing those parts and a third to placing them on palettes. Each of these processes can actually be broken down into sub-tasks, some of which are processing, which must be performed at exact speeds to ensure consistent quality, and others are travel, the machine moving from the end of one processing sub-task to the start of the next. The duration of these latter phases can be manipulated without impacting output as long as the complete task remains within a fixed cycle time to ensure synchronicity with the rest of the production line. 

Given the increasing dependence on sensor data to make real-time adjustments to robotic toolpaths in Industry 4.0, making these adjustments to ensure consistent data transfer may be critical to ensuring quality and avoiding defects in production, i.e. if our welding robot deviates from the seam by 0.1 mm because data from its seam tracker was delayed arriving for processing that could cause pockets and potential joint failure.

In Figure~\ref{fig:tasks}, a caricatured sub-task breakdown for welding could resemble; part localisation, then weld and quality assurance (QA) repeated for each junction. Throughout the  process certain data will be monitored constantly e.g., robot joint positions, TCP position, travel speed etc. (nominally $50$ - $100$ Kbps) but other data requirements are introduced for each sub-task. During each weld pass the robot will move relatively slowly and use a seam tracker to ensure the welding torch follows the junction between two parts exactly. During this sub-task, the seam tracker will measure 200 points and the welding equipment will log current, voltage, feed rate etc. at approximately $200$ Hz. This equates to an additional $650$ Kbps of data being sent to analysis and logging before being used to adapt the robot’s toolpath. During the subsequent QA sub-task, the seam tracking sensor will take many more measurements, $2048$ points at $2$ kHz, approximately $65$ Mbps, one order of magnitude more than during welding. 
At the other end of the spectrum is the palletiser which may be moving for 5s to pack a part, during which intensive data transfer ($65$ Mbps) is required for computer vision but there is a $30$ - $60$s window in which the robot can actually undertake its task before the next part arrives ready for packing. If this process runs concurrently with a QA pass, we need to handle $130$ Mbps whilst if it runs during a welding pass our peak data rate is $66$ Mbps. Scaling up from this simple pair of cells, when many of these, or similar, processes are run concurrently, intentionally misaligning the most data intensive tasks allows for more consistent network performance, leading to improved process performance, as well as less costly networking infrastructure.

%% file: section/model.tex
\section{Industrial Networks: system Model and Optimization Framework}
\label{sec:model}
In this section we provide a generalized system model to capture the aforementioned generic operation of robotic manipulators in an industrial environment. Then, based on the proposed system model a novel optimization framework is proposed to provide an efficient scheduling of the different tasks of the robotic manipulators.

A common scenario in smart factories is that many robots work in a production workshop and each robot operates a sequence of tasks/jobs\footnote{The term \textit{task} and \textit{job} are used interchangeably in the rest of paper.} periodically. Due to the local capacity limitation, the robot needs to communicate with edge cloud during the task. Furthermore, the communication link is shared by all robots. Obviously, massive concurrent communication tasks pose a daunting challenge to the network capacity.  Therefore, on an industrial network with a large number of robotic manipulators transmitting and receiving large amounts of data, without considering the farsighted network evolution, traffic can vary significantly, creating congestion episode that can affect the overall production operation. The aim of this paper is orchestrating the tasks' operating time slot for the different robotic manipulators so that spikes on the transported traffic are avoided having, as a net result, better utilization of network resources.

\begin{figure}[htbp]
    \centering
    \includegraphics[width=.48\textwidth]{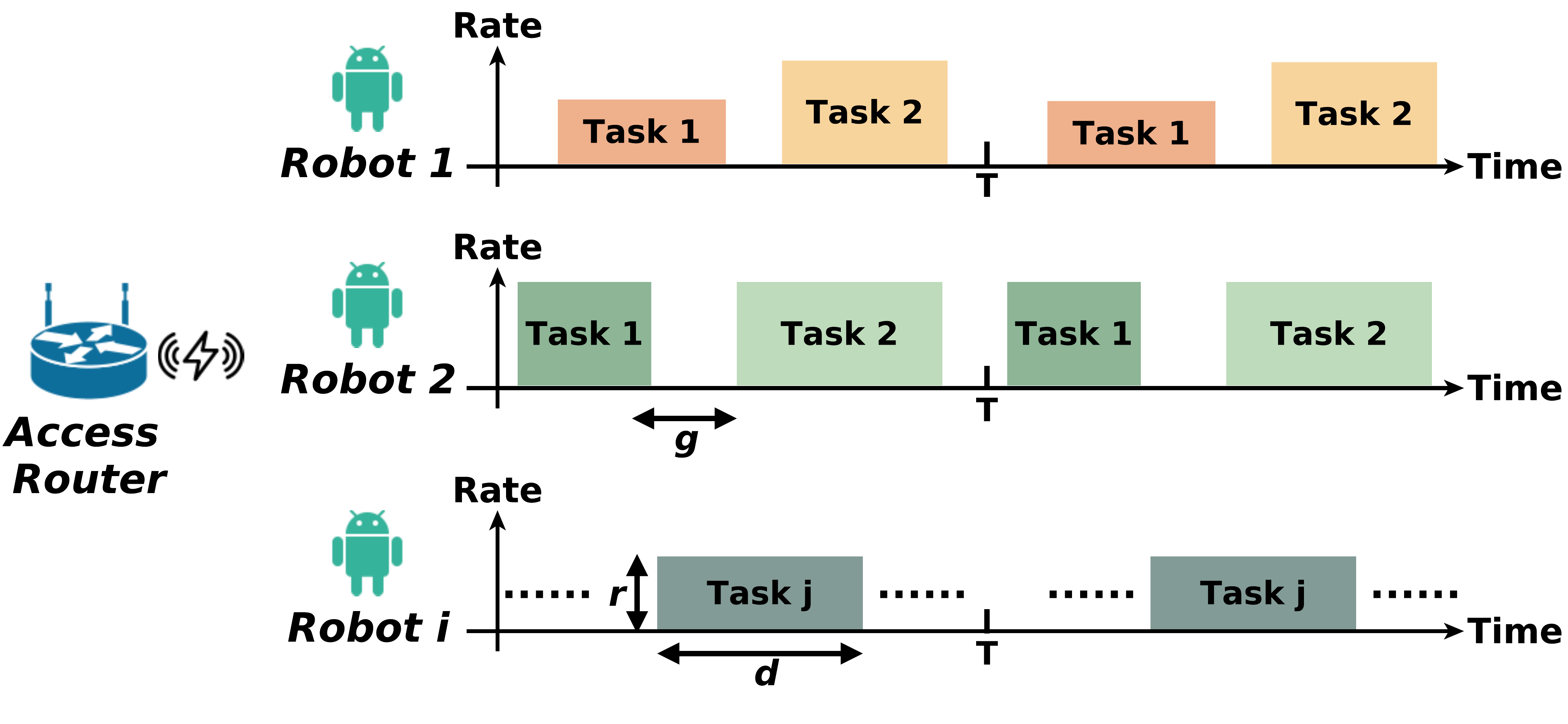}
    \caption{Illustration of a generic robotic manipulators communication model that capture the different tasks and task duration per manipulator.}
    \label{fig:model}
\end{figure}

We use Figure \ref{fig:model} to explain the definition and symbols of the robot communication model. In processing period $T$, we denote the robot set as $\mathcal{I}=\{1,2,\cdots,i,\cdots,I\}$ and each robot deals with a task sequence $\mathcal{J}=\{1,2,\cdots,j,\cdots,J\}$. Therefore, the pair $(i,j)$ locates the $j$th task in robot $i$. Additionally, we use $r_{i,j}$ to show the transmission rate requirement; $d_{i,j}$ to indicate the processing duration; $g^{min}_{i,j}$ to represent the minimum time gap between task $j$ and $j+1$; and $g^{max}_{i,j}$ to show the maximum time gap. Particularly, $g^{min}_{i,J}$ limits the minimum time gap from the last task to the end of period.

Then, we introduce two binary decision variables as follows,

\hangafter 1
\hangindent 1em
\noindent
\\$  x_{i,j,t}=
\begin{cases}
	1,  &\text{the task $(i,j)$ starts from time slot $t$;}  \\
	0,  &\text{otherwise.}
\end{cases}$
\\
\\$ y_{i,j,t}=
\begin{cases}
	1, &\text{the task $(i,j)$ occupies time slot $t$;} \\
	0, &\text{otherwise.}
\end{cases}$
\\

As the number of decision variables depends on the total number of time units in the considered time horizon, it is clear that the choices of values for $T$ and time slot unit $t$ are important. The value of $T$ depends on the robotic process, and, hence, the model should closely follow the different tasks in terms of time durations of the robotic manipulator and the underlying sub-flows. The selection of the time unit $t$ should be such that it allows adequate resolution of the sub-flows by conserving the underlying time structure of the traffic. Furthermore, the complexity of proposed model grows with increased values of $T$ and higher resolution in terms of the time unit $t$. Therefore, a balance should be attained so that the model can provide bounds on the performance (optimal traffic scheduling) for practical, real world applications.
Moreover, to simplify the model, we assume that all robots have the same processing period $T$. Regarding the in-equivalent case, we can set $T$ as the least common multiple among all periods.  

Based on the aforementioned definition, the starting time slot ($\theta^S_{i,j}$) for task $(i,j)$ can be expressed as follows,
\begin{equation}
\label{fml:ST}
\theta^S_{i,j}\!=\!\sum_{t=1}^{T}t\!\cdot\!x_{i,j,t},\forall i\!\in\!\mathcal{I}, j\!\in\!\mathcal{J},
\end{equation}
and the finishing time slot ($\theta^F_{i,j}$) is represented by
\begin{equation}
\label{fml:FT}
\theta^F_{i,j}\!=\!\sum_{t=1}^{T}(t\!+\!d_{i,j}\!-\!1)\!\cdot\!x_{i,j,t}, \forall i\!\in\!\mathcal{I}, j\!\in\!\mathcal{J}.
\end{equation}
Therefore, for the next job $(i,j+1)$, the earliest starting time slot ($\theta^{ES}_{i,j+1}$) depends on the previous task finishing time slot $\theta^F_{i,j}$ and the minimum time gap $g^{min}_{i,j}$ as
\begin{equation}
\label{fml:ES}
\theta^{ES}_{i,j+1}\!=\!\theta^F_{i,j}\!+\!g^{min}_{i,j}\!+\!1\!=\!\sum_{t=1}^{T}(t\!+\!d_{i,j})\!\cdot\!x_{i,j,t}\!+\!g^{min}_{i,j},\forall i\!\in\!\mathcal{I}, j\!\in\!\mathcal{J}\!\setminus\!J.
\end{equation}
Similarly, regarding the latest starting time slot ($\theta^{LS}_{i,j+1}$) of the next job we have
\begin{equation}
\label{fml:LS}
\theta^{LS}_{i,j+1}\!=\!\theta^F_{i,j}\!+\!g^{max}_{i,j}\!+\!1\!=\!\sum_{t=1}^{T}(t\!+\!d_{i,j})\!\cdot\!x_{i,j,t}\!+\!g^{max}_{i,j},\forall i\!\in\!\mathcal{I}, j\!\in\!\mathcal{J}\!\setminus\!J.
\end{equation}


The central aim of this paper is to decrease the peak data rate in the communication link, and the scheduling problem is formulated as follows:	
\begin{subequations}
\label{fml:MLT}
	\begin{align}
	\label{fml:obj}
	&\mathop{\min}_{x_{i,j,t},y_{i,j,t}}\max\sum^{I}_{i=1}\sum^{J}_{j=1} r_{i,j}\!\cdot\!y_{i,j,t},\\
	\label{fml:con1}
	\textrm{s.t.}\quad & \theta^{ES}_{i,j+1}\!\leq\!\theta^S_{i,j+1}\!\leq\!\theta^{LS}_{i,j+1}, \forall i\!\in\!\mathcal{I},j\!\in\!\mathcal{J}\!\setminus\!J,\\
	\label{fml:con2}
	& \theta^F_{i,J}\!+\!g^{min}_{i,J}\!\leq\!T, \forall i\!\in\!\mathcal{I},\\
	\label{fml:con3}
	& y_{i,j,t'}\!\geq\!x_{i,j,t}, \forall i\!\in\!\mathcal{I},j\!\in\!\mathcal{J},t'\!\in\! [t,\min\{t\!+\!d_{i,j}\!-\!1,T\}],\\
	\label{fml:con4}
	& \sum_{t=1}^{T}x_{i,j,t}\!=\!1, \forall i\!\in\!\mathcal{I},j\!\in\!\mathcal{J},\\
	\label{fml:con5}
	& \sum_{t=1}^{T}y_{i,j,t}\!=\!d_{i,j}, \forall i\!\in\!\mathcal{I},j\!\in\!\mathcal{J},\\
	\label{fml:con6}
	& x_{i,j,t},y_{i,j,t}\!\in\!\{0,1\}, \forall i\!\in\!\mathcal{I},j\!\in\!\mathcal{J},t\!\in\![1,T].
	\end{align}
\end{subequations} 
In the objective function \eqref{fml:obj}, $r_{i,j}$ is the transmission data rate and $y_{i,j,t}$ represents the transmission duration. Therefore, $\sum^{I}_i\sum^{J}_j r_{i,j}\cdot y_{i,j,t}$ expresses the total transmission data on time slot $t$. Constraint \eqref{fml:con1} gives the bounds of starting time slot for job $(i,j+1)$. A special case is considered in \eqref{fml:con2} which guarantees the last job should be done within this period and keeps the minimum duration gap. Constraint \eqref{fml:con3} bridges two decision variables, $x_{i,j,t}$ and $y_{i,j,t}$, and enforces that if the job starts form time slot $t$ then the next $d_{i,j}$ durations should also be set for processing/occupation purpose. Especially, the upper bound of the duration should be within the same period, i.e. $\min\{t+d_{i,j}-1,T\}$. \eqref{fml:con4} guarantees all jobs should be scheduled and only once. In the end, \eqref{fml:con5} promises that for task $(i,j)$, the number of occupied time slots should be exactly match the processing durations. 

To linearize the optimization model \eqref{fml:MLT}, we introduce an auxiliary variable $z$ which is defined as
\begin{equation}
	z\!=\!\max\sum^{I}_{i=1}\sum^{J}_{j=1} r_{i,j}\!\cdot\!y_{i,j,t}. \nonumber
\end{equation}
Therefore, \eqref{fml:MLT} can be rewritten in the form of integer linear programming (ILP) model:
\begin{subequations}
\label{fml:ILP}
	\begin{align}
	\label{fml:obj_new}
	&\mathop{\min}_{x_{i,j,t},y_{i,j,t},z}z\\
	\label{fml:con1_new}
	\textrm{s.t.}\quad
	& \sum_{i=1}^I\sum_{j=1}^J r_{i,j}\!\cdot\!y_{i,j,t}\!\leq\!z, \forall t\!\in\![1,T]\\
	& \eqref{fml:con1}\sim\eqref{fml:con6} \nonumber
	\end{align}
\end{subequations}

%% file: section/heuristic.tex
\section{A Heuristic Algorithm for Real-Time Scheduling}
\label{sec:heuristic}

Generally, solving ILP models is very time-consuming. By reviewing the optimization model, we notice that constraint \eqref{fml:con1} gives the earliest and latest starting time for jobs $\{2,\cdots,J\}$, and constraint \eqref{fml:con2} provides the latest starting time for job $J$. In other words, for the last job $J$, the latest starting time $\theta^{LS}_{i,J}$ is limited by \eqref{fml:con1} and \eqref{fml:con2} together. Once $\theta^{LS}_{i,J}$ is determined, $\theta^{LS}_{i,J-1}$ is also influenced by \eqref{fml:con2} indirectly to leave enough time slots for last job $J$ to be done within time period $T$. Thus, we have two latest starting time limitations for the job $j'\in\{2,\dots,J\}$, which are computed as
\begin{subequations}
\label{fml:LS_def}
\begin{align}
    \label{fml:LS_def1}
    & \theta^{LS1}_{i,j'}\!=\!\theta^F_{i,j'\!-\!1}\!+\!g_{i,j'\!-\!1}^{max}\!+\!1, &&\forall i\!\in\!\mathcal{I},\\
    \label{fml:LS_def2}
    & \theta^{LS2}_{i,j'}\!=\!T\!-\!\sum_{j=j'}^{J}g_{i,j}^{min}\!-\!\sum_{j=j'}^{J}d_{i,j}\!+\!1, && \forall i\!\in\!\mathcal{I},
\end{align}
\end{subequations}
respectively. Then we select the minimum as the latest starting time for job $j'$, Once both boundaries ($\theta^{ES}_{i,j}$ and $\theta^{LS}_{i,j}$) are determined, one proper method is allocating each job in this area randomly. Then we rerun this process in several iterations and pick the minimal peak data rate assignment as output. The proposed randomized time window preservation algorithm (RTWPA) is detailed in Algorithm \ref{alg:RTWPA}.
\begin{algorithm}
    \caption{Randomized Time Window Preservation Algorithm (RTWPA)}
    \label{alg:RTWPA}
	\KwData{the number of loops $N$; robots set $\mathcal{I}$; jobs set $\mathcal{J}$; transmission rate requirement $r_{i,j}$; processing duration requirement $d_{i,j}$; minimum time gap $g^{min}_{i,j}$; maximum time gap $g^{max}_{i,j}$}
	\KwResult{time slot allocation $y_{i,j,t}$}
	\While{not reach $N$ loops}{
	    	\ForEach{robot $i$ in $\mathcal{I}$, task $j$ in $\mathcal{J}$}{
	    	    \eIf{$j==1$}{
	    	        $\theta^{ES}_{i,j}\leftarrow1$\;
	    	        $\theta^{LS}_{i,j}\leftarrow T\!-\!\sum_{j=1}^J g^{min}_{i,j}\!-\!\sum_{j=1}^J d_{i,j}\!+\!1$\;
	    	    }
	    	    {
	    	        $\theta^{ES}_{i,j}\leftarrow\theta^F_{i,j-1}\!+\!g^{min}_{i,j}\!+\!1$\;
	    	        calculate $\theta^{LS1}_{i,j}$ and $\theta^{LS2}_{i,j}$ via \eqref{fml:LS_def1} and \eqref{fml:LS_def2} respectively\;
	    	        $\theta^{LS}_{i,j}\leftarrow\min(\theta^{LS1}_{i,j},\theta^{LS2}_{i,j})$\;
	    	    }
	        	\If{$\theta^{ES}_{i,j}>\theta^{LS}_{i,j}$}{ 
	        		throw an error and return\;
	        	}
	        	$\theta^S_{i,j}\leftarrow[\theta^{ES}_{i,j},\theta^{LS}_{i,j}]$\tcp*{choose the starting time randomly}
	        	$\theta^F_{i,j}\leftarrow\theta^S_{i,j}\!+\!d_{i,j}-1$\;
	        	$y_{i,j,\theta^S_{i,j}},\cdots,y_{i,j,\theta^F_{i,j}}\leftarrow1$\;
	        }
	    record the allocation $y_{i,j,t}$\;
	    calculate $z=\max\sum_i\sum_j r_{i,j}\cdot y_{i,j,t}$\;
	}
	pick the minimal $z$ and output $y_{i,j,t}$ accordingly\; 
\end{algorithm}

There are three loops in Algorithm \ref{alg:RTWPA}, and, thus, the time complexity of RTWPA is $O(N\!\times\!I\!\times\!J)$. In particular, $N$ is a hyperparameter which controls the number of random search iterations. More iterations results in better performance but longer computation time. Based on the parameter settings in Section \ref{sec:numerical} and after several rounds of grid search, for our simulation scenarios the best configuration recorded is $N=10^3$. Based on the above, the time complexity of RTWPA becomes $O(I\!\times\!J)$.

%% file: section/numerical.tex
\section{Numerical Investigations}
\label{sec:numerical}

In this section, we demonstrate the performance of the optimal decision making derived form the proposed ILP model \eqref{fml:ILP} together with the defined heuristic algorithm RTWPA that allows real-time decision making. To further benchmark the results we also compare with the nominal case where robotic manipulators are generating traffic in an uncoordinated manner; this case is named as the No-Scheduling regime in the discussion hereafter. The comparison is considered under various scenarios with different robot and task combinations. In particular, we set the number of tasks per robot as 1, 2, and 3, which correspond to the light-, medium-, and heavy-duty cases.
The simulation parameters are summarized in Table \ref{tab:Parameters}. 
	
\addtolength{\topmargin}{0.03in}
\begin{table}[htbp]
	\centering
	\caption{Summary of Simulation Parameters.}
	\label{tab:Parameters}
    \small{\singlespacing{
	\begin{tabular}{cl}
		\toprule
		\textbf{Parameter} & \textbf{Value} \\
		\midrule
		robots set ($\mathcal{I}$) & \{2,4,6,8,10\} \\
		tasks set ($\mathcal{J}$) & \{1,2,3\} \\
		time period ($T$) & 15 \\
		transmission rate ($r_{i,j}$) & [1,4] Mbps\\
		processing slot ($d_{i,j}$) & [1,3] \\
		minimum gap ($g^{min}_{i,j}$) & 1 \\
		maximum gap ($g^{max}_{i,j}$) & [3,5] \\
		\bottomrule
	\end{tabular}
	}}
\end{table}

\begin{figure}[htbp]
    \centering
    \includegraphics[width=\figSize\textwidth]{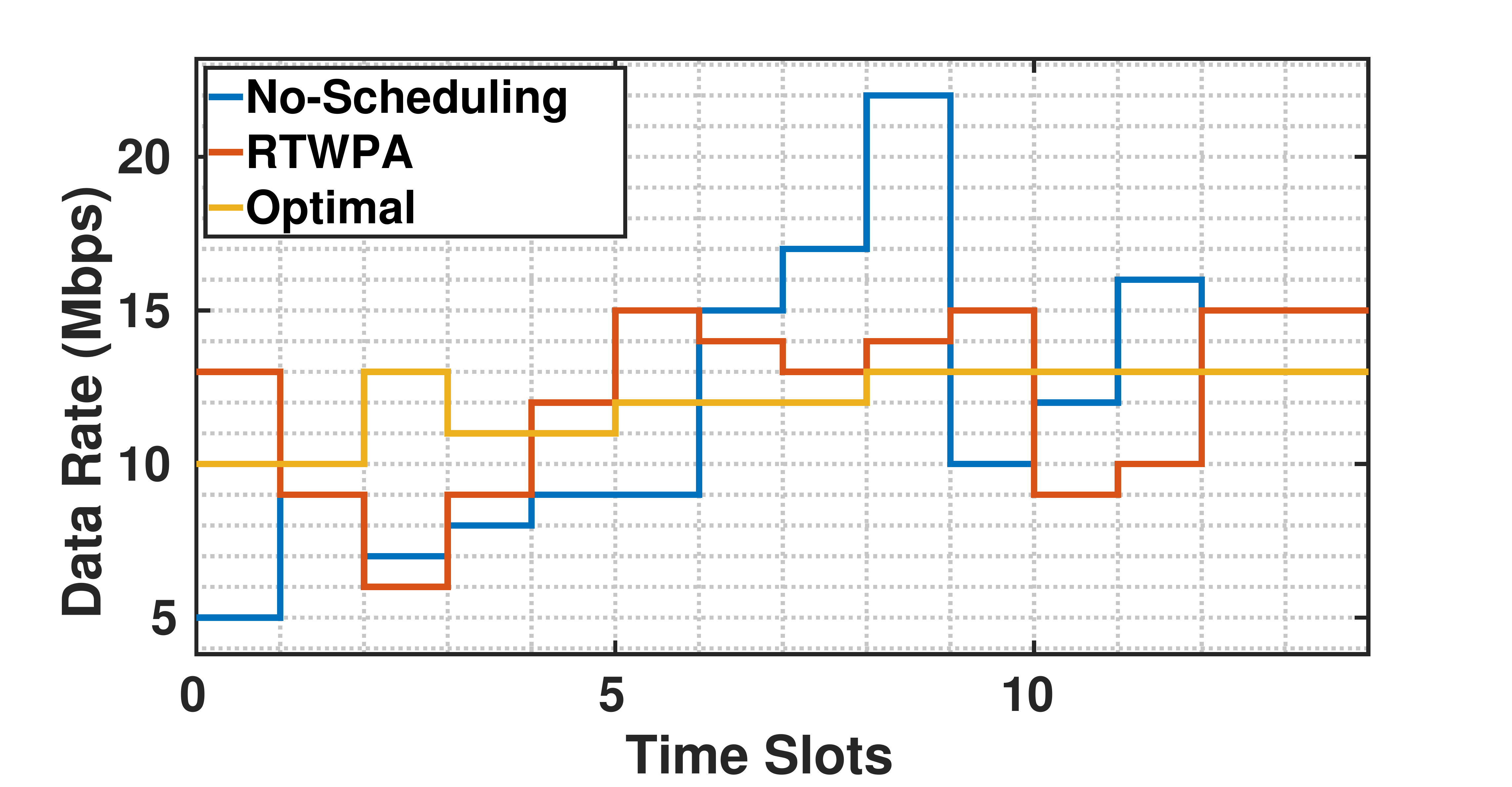}
    \caption{Data rate under a heavy-duty case ($I\!=\!10,J\!=\!3$).}
    \label{fig:task3case}
\end{figure}

Figure~\ref{fig:task3case} presents the data rate distributions for different allocation approaches under a heavy-duty case. We can see that the optimal solution, which is generated by our ILP model, holds steady during the process (between $10$ and $12$ Mbps) and can provide a smooth communication; while the proposed heuristic algorithm RTWPA, whose value varies from $6$ to $15$ Mbps, contains some fluctuations; the data rate of No-Scheduling changes heavily from $5$ to $22$ Mbps. From the results, we can verify that the proposed methods can be applied to flatten traffic communications. 

\begin{figure}
\centering
    \begin{subfigure}[b]{\figSize\textwidth}
        \includegraphics[scale=\figScale]{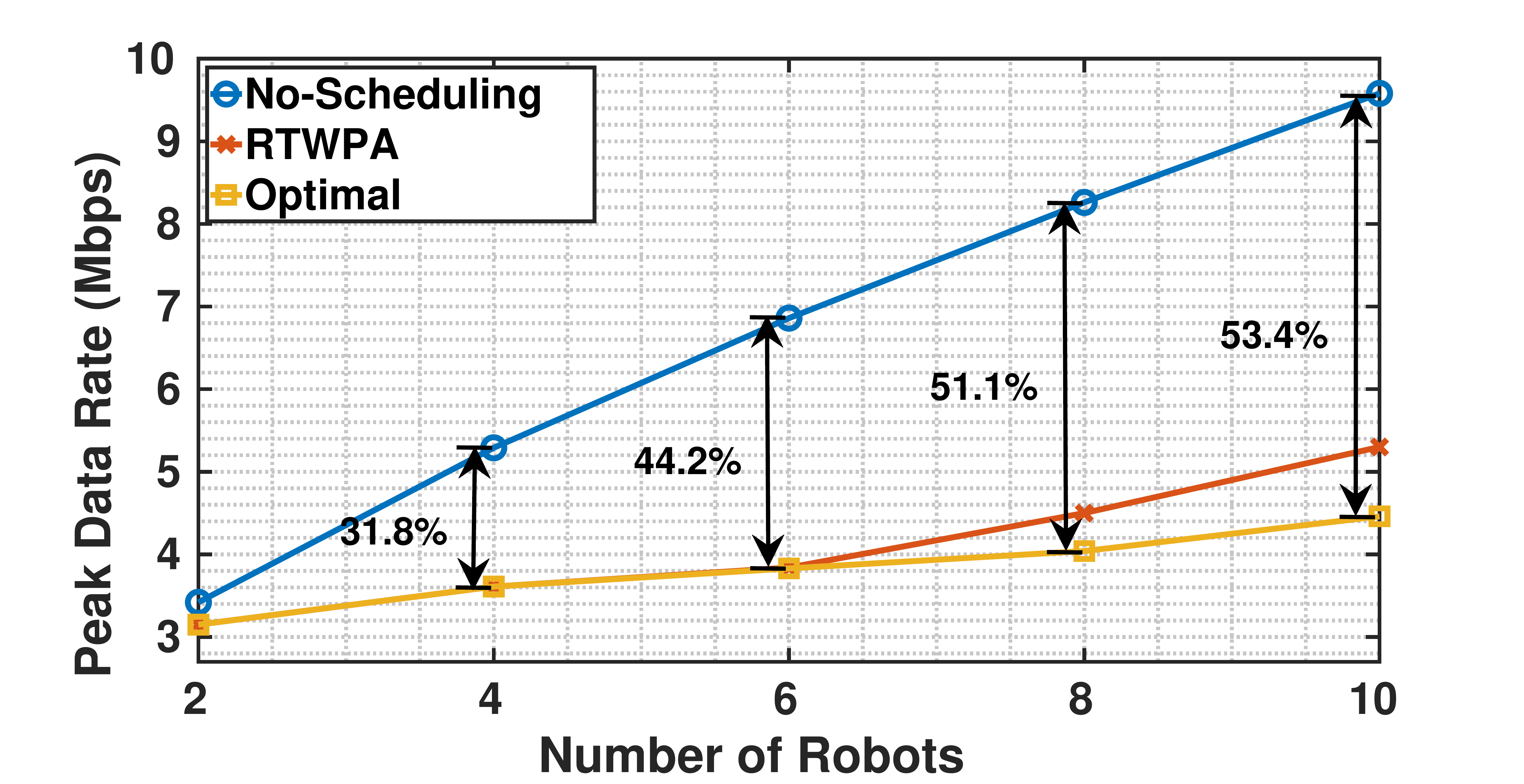}
        \caption{Light-duty case ($J\!=\!1$).}
        \label{fig:task1}
    \end{subfigure}
    \begin{subfigure}[b]{\figSize\textwidth}
        \includegraphics[scale=\figScale]{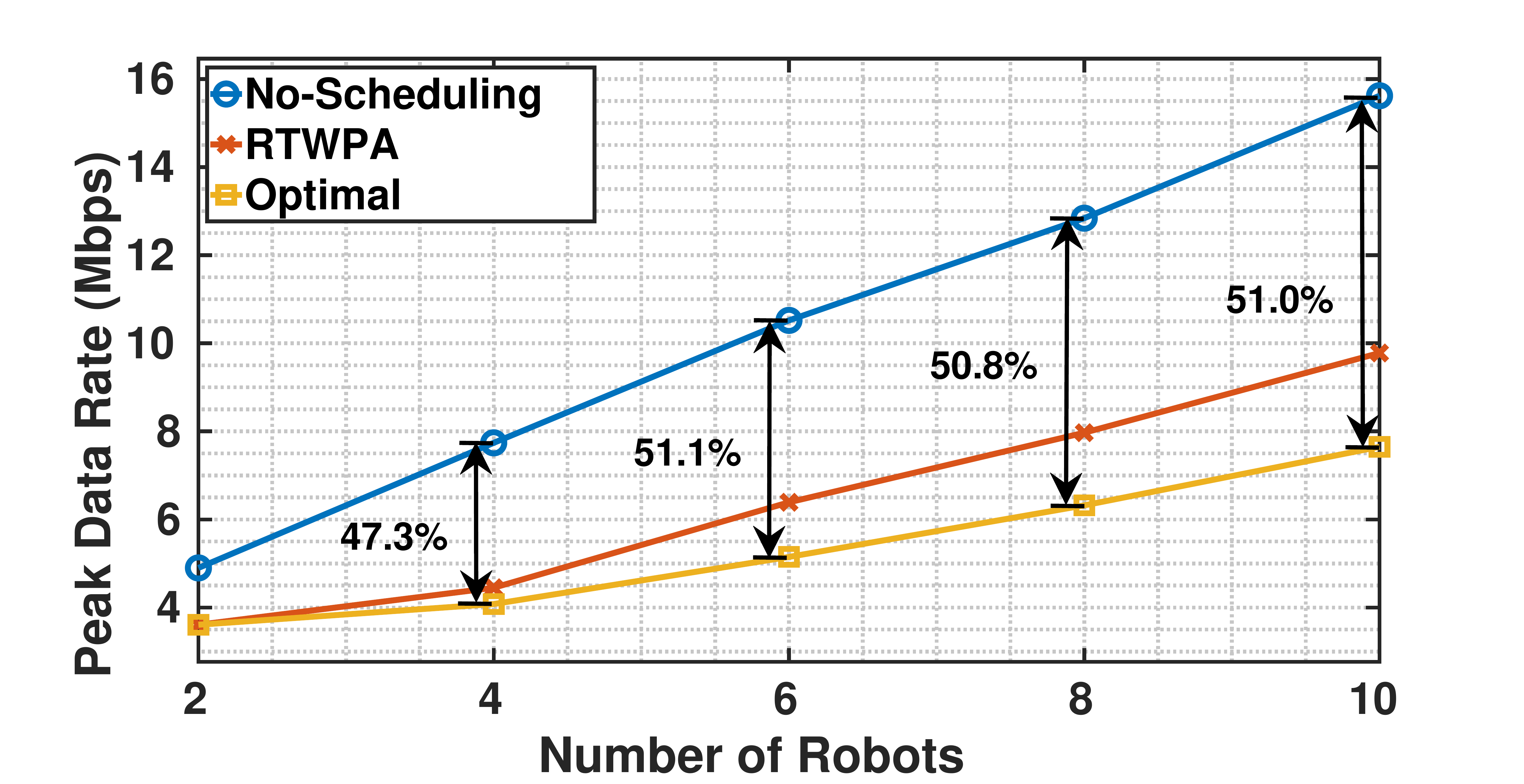}
        \caption{Medium-duty case ($J\!=\!2$).}
        \label{fig:task2}
    \end{subfigure}
    \begin{subfigure}[b]{\figSize\textwidth}
        \includegraphics[scale=\figScale]{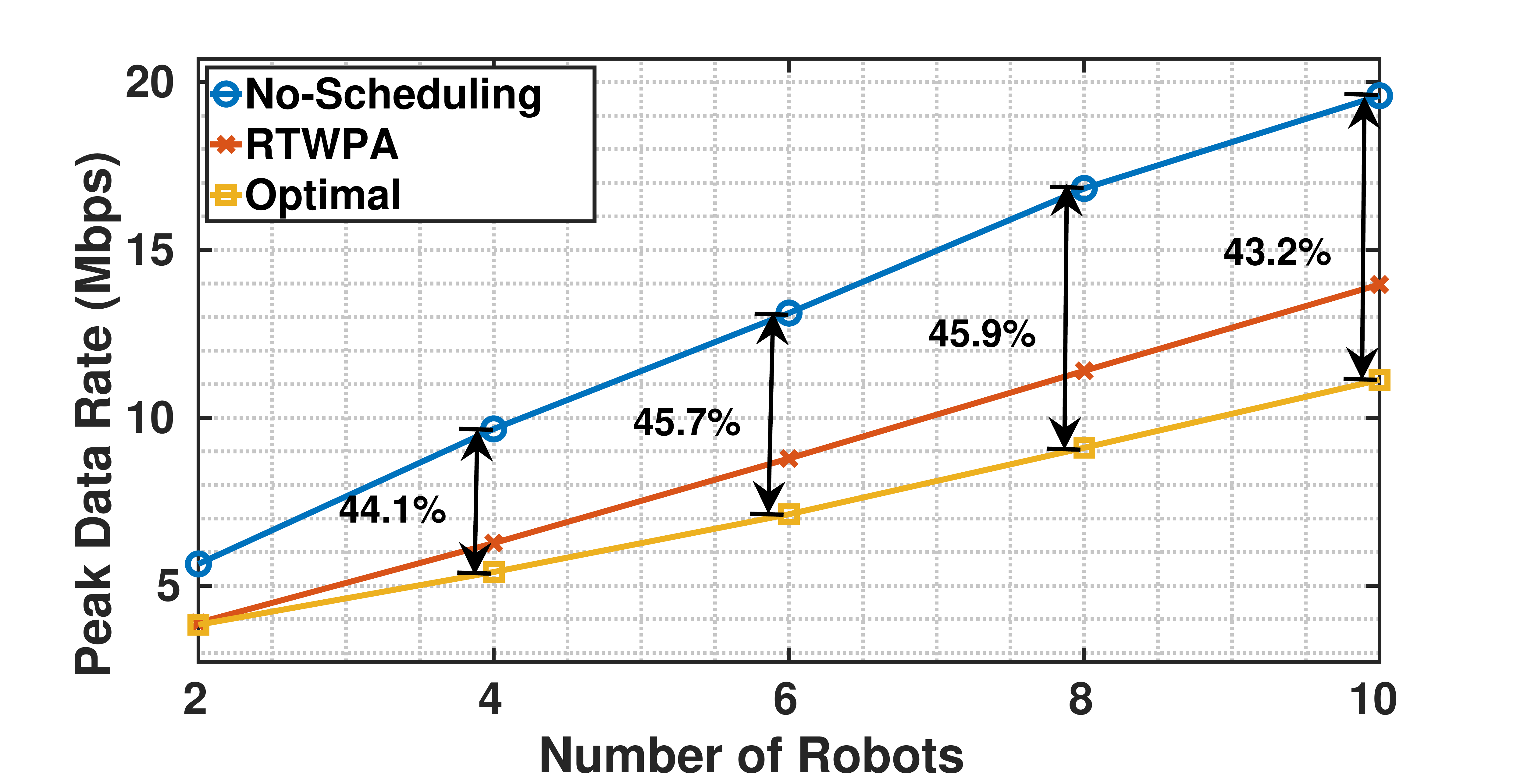}
        \caption{Heavy-duty case ($J\!=\!3$).}
        \label{fig:task3}
    \end{subfigure}
    \caption{Peak data rate comparison.}
    \label{fig:rate}
\end{figure}

In Figure~\ref{fig:rate}, we evaluate the peak data rates of different algorithms, which are averaged under 100 various simulation scenarios. In general, the proposed ILP model and the heuristic algorithm RTWPA can provide better traffic allocations. We also label the percent of performance reduction (PoPR) for the optimal solutions in Figure~\ref{fig:rate}, which is defined as the ratio of traffic reduced rate compared with No-Scheduling. 

In Figure~\ref{fig:task1}, which relates to the light-duty case, the achieved PoPR, with respect to the case of no coordination, shows a steady increase with the growth of the number of robotic manipulators from $2$ to $10$. However, as can be observed in Figure~\ref{fig:task2} for the medium-duty case, the achieved PoPR, with respect to the case of no coordination, has a rise from $2$ to $6$ robots and then remains steady after that. Furthermore, Figure~\ref{fig:task3} which relates to the heavy-duty case reveals that the PoPR percentage gain, with respect to no coordination, saturates with a small decline in value from $8$ to $10$ robots. What is interesting is that the trend of PoPR performs similarly when we compare in the horizontal. For example, when the number of robots is $6$, the PoPR has a rise from $44.2\%$ (Figure~\ref{fig:task1}) to $51.1\%$ (Figure~\ref{fig:task2}) then drops to $45.7\%$ (Figure~\ref{fig:task3}). This pattern also holds on RTWPA. The reason is that the PoPR is affected by the number of robots ($I$) and the number of tasks ($J$) combinationally. When the value of $I$ or $J$ is not large, all methods, including the ILP model, RTWPA, even No-Scheduling, can find an allocation to avoid traffic congestion. Thus, the performance gaps among these three approaches are inconspicuous. Then, with the growth of $I$ or $J$, the proposed methods can find better allocations by placing task blocks in the proper time slots, and the PoPR increases accordingly. However, if $I$ or $J$ is large enough, the traffic congestion is inevitable and even the ILP model also suffers from the stacked peak data rate. Therefore, the PoPR becomes saturated even decreased. It can be expected that in the extreme case, when $J$ is fairly high and squeezes $\theta^{ES}\!=\!\theta^{LS}$, all three approaches produce the exact same allocation then PoPR becomes $0$.  

\addtolength{\topmargin}{0.03in}
\begin{table*}
\centering
\caption{The Average Running Time Comparison}
\label{tab:running_time}
\small{\singlespacing{
\begin{tabular}{c|c|c|c|c|c|c|c}
\hline
\multicolumn{3}{c|}{\multirow{2}{*}{}} & \multicolumn{5}{c}{ Number of Robots ($I$)} \\ \cline{4-8} 
\multicolumn{3}{c|}{} & 2 & 4 & 6 & 8 & 10 \\ \hline
\multirow{9}{*}{\begin{tabular}[c]{@{}c@{}}  Number\\ of Tasks ($J$)\end{tabular}} & \multirow{3}{*}{No-Coordination} & 1 & 0.006ms & 0.010ms & 0.013ms & 0.016ms & 0.019ms \\ \cline{3-8} 
 &  & 2 & 0.021ms & 0.039ms & 0.058ms & 0.075ms & 0.095ms \\ \cline{3-8} 
 &  & 3 & 0.035ms & 0.066ms & 0.097ms & 0.127ms & 0.186ms \\ \cline{2-8} 
 & \multirow{3}{*}{RTWPA} & 1 & 6.540ms & 9.788ms & 12.988ms & 16.188ms & 19.496ms \\ \cline{3-8} 
 &  & 2 & 21.479ms & 38.768ms & 57.157ms & 75.418ms & 92.844ms \\ \cline{3-8} 
 &  & 3 & 35.025ms & 66.707ms & 98.061ms & 0.126s & 0.154s \\ \cline{2-8} 
 & \multirow{3}{*}{Optimal} & 1 & 0.114s & 0.331s & 0.685s & 8.458s & 150.374s \\ \cline{3-8} 
 &  & 2 & 0.362s & 1.292s & 11.384s & 171.644s & 39.912min \\ \cline{3-8} 
 &  & 3 & 0.319s & 2.549s & 13.680s & 201.099s & 20.213min \\ \hline
\end{tabular}
}}
\end{table*}

Table~\ref{tab:running_time} illustrates the computation complexities of No-Scheduling, RTWPA and the optimal solution\footnote{To put computational times in perspective we note that the simulation is on MATLAB R2020b in a 64-bit Ubuntu 20.04.01 LTS environment on a machine equipped with an Intel Core i5-650 CPU 3.20GHz and 8GB RAM.}. The performance gains of the proposed ILP model in peak data rate is at the expense of significant larger running times than the RTWPA algorithm. As can be seen in Table~\ref{tab:running_time}, the running time of No-Scheduling and RTWPA rise with the increase of robots and tasks. Additionally, the computation time of the optimal is increasing with the growth of $I$. It is worth noting that when $I\!=\!10$, $J\!=\!2$, the ILP model needs around 40 minutes to calculate optimal allocations; while for $J\!=\!3$, the computation time is approximately halved. This is because the searching space is a dominant factor for the running time. Taken the decision variable $x_{i,j,t}$ into consideration, on the one hand, the increased number of robots and tasks correspond to the extension in $i$ and $j$ dimension respectively, which expands the searching space; on the other hand, immense number of tasks would cause supersaturation, and, then, squeeze the gap between $\theta^{ES}$ and $\theta^{LS}$. As a result, the searching space is also reduced since $\theta^{ES}$ and $\theta^{LS}$ determine the boundary of constraint~\eqref{fml:con1} in ILP model.

%% file: section/conclusions.tex
\section{Conclusions}
\label{sec:concl}
Undoubtedly, the integration of 5G technologies and proliferation of industrial robots can be deemed as the key drivers towards the Industry 4.0 vision. In this emerging ecosystem in this paper the central aim is to reduce peak traffic load in an industrial edge cloud supported 5G network with a large number of robotic manipulators. Reducing the peak traffic demand in the industrial network would allow for network slicing solutions that require less slack bandwidth to hedge against uncertain increases in the volume of traffic per specific slice. In addition, reducing peak aggregate traffic reduce queuing length and hence better latency which is important in several time-sensitive robotic manipulator actions. A wide set of numerical investigations reveal that by capitalizing on the pseudo deterministic traffic profile of industrial robots peak traffic can be reduced by more than 50\% compared to the nominal case where traffic originated from robotic manipulators allowed to traverse the industrial network without coordination.  